\documentclass[10pt,twocolumn,letterpaper]{article}

\usepackage{iccv}              %
\definecolor{iccvblue}{rgb}{0.21,0.49,0.74}
\usepackage[pagebackref,breaklinks,colorlinks,allcolors=iccvblue]{hyperref}
\usepackage[accsupp]{axessibility}
\usepackage{booktabs}
\usepackage[normalem]{ulem}
\useunder{\uline}{\ul}{}

\def\paperID{237} %
\def\confName{ICCV}
\def\confYear{2025}

\title{MoMaps: Semantics-Aware Scene Motion Generation with Motion Maps}

\author{
        Jiahui Lei\textsuperscript{1,2} \quad Kyle Genova\textsuperscript{1} \quad  George Kopanas\textsuperscript{3} \quad Noah Snavely\textsuperscript{1} \quad Leonidas Guibas\textsuperscript{1}\\
        $^1$ Google DeepMind \qquad
        $^2$ University of Pennsylvania \qquad $^3$ Google\\
        {\tt\small leijh@cis.upenn.edu, \{jiahuilei,kgenova,gkopanas,snavely,guibas\}@google.com} 
    }

\begin{document}
\twocolumn[\maketitle\vspace{-3em}
\begin{center}
\includegraphics[width=1.0\linewidth]{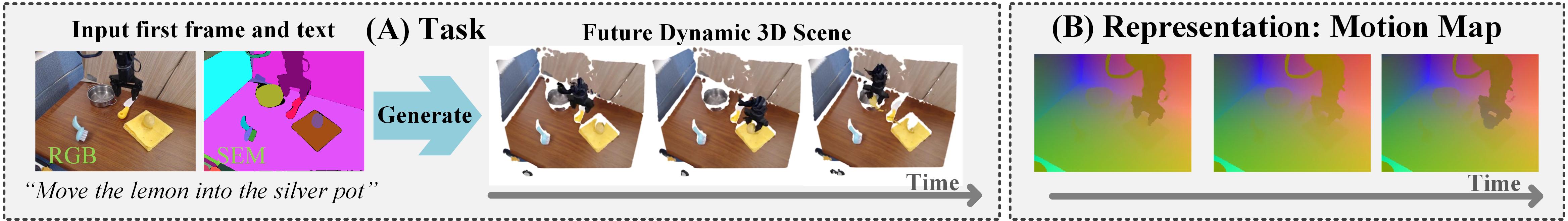}
\captionof{figure}{(A) Given the first time frame color and segmentation image and a text prompt, our model generates the future dynamic 3D scene. (B) 3D dynamic scenes as pixel-aligned curve/trajectory images, re-purposing an image diffusion model for 4D generation. 
}
\label{fig:teaser}
\end{center}
\vspace{-2em}
\bigbreak]
\vspace{-2em}
\begin{abstract} 
This paper addresses the challenge of learning semantically and functionally meaningful 3D motion priors from real-world videos, in order to enable 
prediction of future 3D scene motion from a single input image.
We propose a novel pixel-aligned \textit{Motion Map} (MoMap) representation for 3D scene motion, which can be generated from existing generative image models to facilitate efficient and effective motion prediction. 
To learn meaningful distributions over motion, we create a large-scale database of MoMaps from over 50,000 real videos and train a diffusion model on these representations. 
Our motion generation not only synthesizes trajectories in 3D but also suggests a new pipeline for 2D video synthesis: first generate a MoMap, then warp an image accordingly and complete the warped point-based renderings. 
Experimental results demonstrate that our approach generates plausible and semantically consistent 3D scene motion. \end{abstract} 
\vspace{-2em}
    
\section{Introduction}
\label{sec:intro}
Motion is ubiquitous in the visual world. 
Understanding, reconstructing, forecasting, and interrogating how objects and entities move in 3D is an important task for computer vision and is critical for applications that involve interaction with physical environments such as augmented reality, autonomous driving, or robotics. 
Of special interest to us is motion forecasting, prediction, and generation, as required by an agent operating in a 3D environment. 
Despite its importance, there is a lack of approaches that learn 3D generative motion priors at scale. 
Existing methods either treat dynamic scene priors as a 2D video generation problem, focus on localized object-centric priors from large-scale object databases, or handle only very short 3D trajectories in constrained, small-scale setups.

\begin{figure*}[h]
    \centering
    \includegraphics[width=1.0\linewidth]{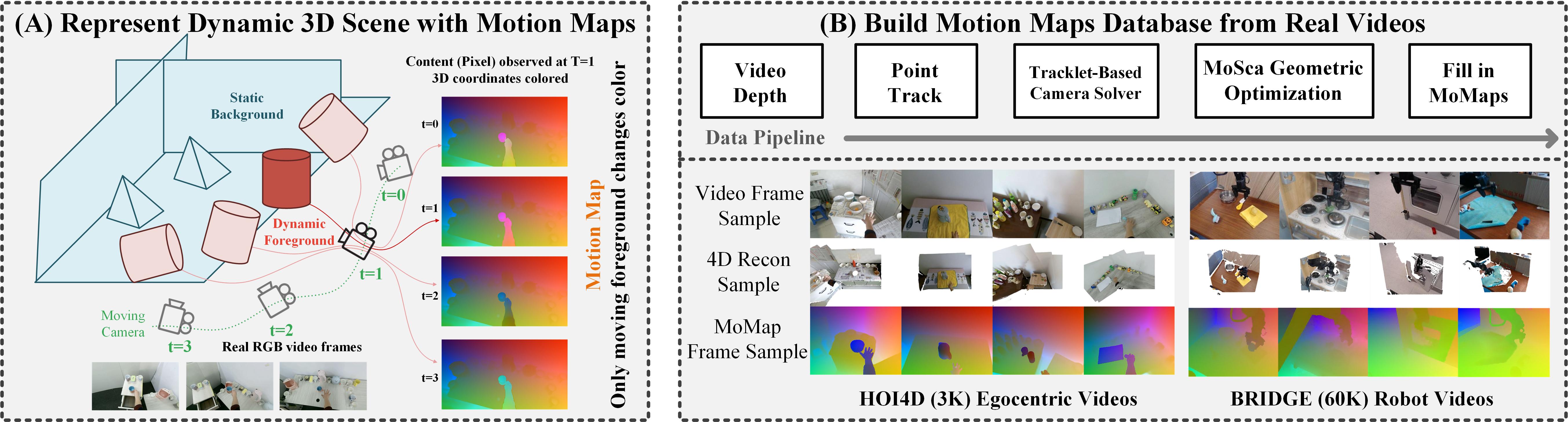}
    \caption{\small\textbf{Motion Maps}: 
    (A) Dynamic 3D scenes can be represented as one or more Motion Maps -- curve/trajectory images. 
    (B) We develop a full-stack data pipeline to recover a large dataset of MoMaps from many real videos.
    }
    \label{fig:momap_repr}
    \vspace{-2em}
\end{figure*}

In this work, we aim to learn semantically and functionally meaningful 3D motion priors from real-world videos at scale, and to use these priors for semantics-aware 3D motion generation. 
Recent generative motion methods either consider 2D motions (e.g., TAPIR~\cite{doersch2023tapir}) or treat 3D trajectories as sets of very short curves, using PointNet or Transformer architectures over them (e.g., GeneralFlow~\cite{gflow}, Track2Act~\cite{track2act}). 
In contrast, we represent motion in a video through a set of pixel-aligned 3D motion snippets we call \textbf{Motion Maps}, or \textbf{MoMaps} for short (Fig.~\ref{fig:momap_repr}-A).
Each MoMap is a motion snapshot---a view of the evolving scene from a fixed camera over a time interval (typically about 50 frames at 3 FPS). 
The pixels of each MoMap frame encode (in RGB channels) the XYZ 3D locations of points seen by that camera in the fixed reference camera coordinate system, and the entire MoMap records their spatial evolution during the time interval captured by the MoMap. 

MoMaps facilitate motion learning in two ways. First, they disentangle camera motion from object motion, reducing the dimensionality of the problem (only dynamic foreground changes over time). 
Second, their image-like structure allows us to build upon diffusion-based image generative models, such as Stable Diffusion~\cite{podell2023sdxl,rombach2022high}, that have been trained on vast amounts of data---effectively repurposing them from the task of dense photometric generation to that of dense position and motion prediction, considering the full context (both foreground and background) of the scene. 
Such repurposing has been used, for example, in depth prediction by the Marigold method~\cite{ke2023repurposing}.
Additionally, pixel-aligned motion images can directly bind with powerful 2D segmentation masks such as SAM~\cite{kirillov2023segment,ravi2024sam}, enabling us to exploit semantic and instance information when predicting motion, as motion and semantics are often strongly correlated.

Prior motion prediction methods typically rely on synthetic data (e.g., Kubric~\cite{greff2021kubric}) for training, where the motions involved do not capture the richness of the physical world. 
In contrast, our goal is to learn semantically meaningful real-world motions. 
Leveraging recent advances in point tracking and 4D reconstruction, our work presents a large dataset of 3D motions extracted from real-world videos and represented as MoMaps. 
Towards this goal, we developed a full-stack data pipeline based on video depth models~\cite{hu2024-DepthCrafter,wang2025continuous,video_depth_anything}, 3D point tracking~\cite{ngo2024delta,xiao2024spatialtracker}, video object segmentation~\cite{cheng2023tracking,ravi2024sam} and the MoSca 4D reconstruction system~\cite{lei2024mosca} to generate 3D tracks and semantic maps for more than 50K real videos taken from the \textbf{HOI4D}~\cite{liu2022hoi4d} (egocentric) and \textbf{BRIDGE}~\cite{walke2023bridgedata} (robotics) datasets.

Learning a 3D motion prior from large-scale real videos and building a generative model to predict future 3D trajectories from a single initial frame has proven valuable for various tasks, including robotics~\cite{gflow} and video model prompting~\cite{geng2024motionprompting}. 
As the first work to explore dense 3D trajectory generation, we showcase another novel downstream application: 2D video frame synthesis. 
Once a MoMap is generated, we can render the dense 3D trajectories frame by frame into target cameras, resulting in a partial video. 
A lightweight image diffusion model with flatten-cross-attention can then complete these partially warped 2D frames, directly synthesizing 2D RGB and semantic videos. 
This pipeline highlights a new direction for video generation that explicitly embeds dynamic 3D bias and 3D motion consistency into the video formation process.
Because MoMap retains an image-like structure, most techniques that operate with 2D image diffusion models can be adapted to MoMaps, offering increased flexibility and opening up a broad range of future enhancements for 3D motion generation, for example, more fine-grained motion generation control through vision-language models.

In summary, our key contributions are:
\textbf{(1)} A general framework for representing 3D motion in dynamic scenes that can be used for encoding real videos as well as generating plausible future motions conditioned by scene view semantics and language.
\textbf{(2)} An image-like representation of motions, MoMaps, that disentangles camera and object motions and allows the use of large pre-trained image diffusion models for motion prediction.
\textbf{(3)} A large database of MoMaps derived from open-source datasets. 
\textbf{(4)} An application of MoMaps to a new paradigm for generating future video frames from an image by first generating MoMaps and then completing renderings of warped point clouds.

\section{Related Work}
\label{sec:related}
\paragraph{Motion Generation in 2D and 3D}
The closest related work to ours typically focuses on predicting 2D or 3D point trajectories, often within robotics settings~\cite{gflow,xu2024flow,bharadhwaj2024track2act,wen2023any,niu2025pre}:
TAPIR~\cite{doersch2023tapir} uses a U-Net and Fourier encodings of 2D tracks to predict future 2D trajectories from large-scale real videos.
GeneralFlow~\cite{gflow} adopts a PointNeXT geometric backbone to process a point cloud from the first video frame, and employs a trajectory-conditioned VAE to predict future 3D trajectories for queried object points.
Im2Flow2Act~\cite{xu2024flow} leverages Animatediff~\cite{guo2023animatediff} to predict 2D tracks and visibilities obtained from TAPIR~\cite{doersch2023tapir}, focusing on object manipulation tasks.
Track2Act~\cite{bharadhwaj2024track2act} uses a transformer-based diffusion model over sets of 2D object tracks; it generates these tracks from the first frame and then estimates a rigid SE(3) transformation to manipulate the object. Similarly, ATM~\cite{wen2023any} learns policies on top of a comparable transformer-based diffusion model for 2D track generation.
In a concurrent effort, ARM4R~\cite{niu2025pre} pre-trains a causal transformer to predict future 3D tracks (at a coarser resolution in an egocentric coordinate system) on EPIC-Kitchens, later fine-tuning on robot joint trajectories for policy learning.
Zhang et al.~\cite{zhang2024dynamic} address action-conditioned 3D future prediction for specific non-rigid objects reconstructed from visual data.
In contrast to these works, our model focuses on long-term, dense, and camera-disentangled 3D motion generation. It also exploits priors from pre-trained image diffusion models, extending beyond the robotics domain and offering a more general-purpose solution.

\paragraph{Video Generation}
The field of video generation has advanced rapidly with models such as SORA~\cite{liu2024sora}, Veo~\cite{veo2024}, CogVideo~\cite{yang2024cogvideox}, and COSMOS~\cite{agarwal2025cosmos}, etc., which primarily capture priors over pixel sequences across multiple frames. 
Their approaches often rely on diffusion-based methods applied to stacks of frames~\cite{blattmann2023stable,ma2024latte,lin2024open,ge2023preserve} or autoregressive models that predict future frames~\cite{kondratyuk2023videopoet,liu2024stablev2v,deng2024nova}. 
Some works~\cite{tang2024embodiment,ko2023learning} first generate future 2D video frames and then extract 3D motion for downstream tasks such as object manipulation. 
However, in all these methods, motion priors are implicitly learned through pixel-level color changes rather than explicit 3D trajectories.
By contrast, our approach learns a prior directly on dense 3D tracks, which reside on a simpler and smoother manifold compared to pixel-space motion representations. 
Additionally, recent works~\cite{geng2024motionprompting,bian2025gsdit,gu2025das} demonstrate that 3D tracks reconstructed from reference videos or obtained via user prompts can improve controllability in video generation. 
Yet these methods still rely on reconstructing 3D motion from existing sequences, whereas we focus on generating new 3D motions outright.

\paragraph{4D Reconstruction}
Reconstructing real-world, semantically meaningful 3D motion from video is a key step in learning 3D motion priors. Modern vision foundation models such as video depth estimators~\cite{hu2024-DepthCrafter,wang2025continuous,video_depth_anything}, 2D/3D point trackers~\cite{xiao2024spatialtracker,ngo2024delta,karaev2023cotracker,doersch2023tapir}, and the induced 4D reconstruction systems~\cite{lei2024mosca,zhang2024monst3r,wang2025continuous,badki2025l4p} play a crucial role in this process. They enable camera pose estimation, geometry reconstruction, and the establishment of dense 3D correspondences, which collectively encode the 3D motion present in a scene. These recent advances in 4D reconstruction serve as the foundation for constructing large-scale databases, facilitating the learning of predictive and generative motion priors.

\section{Method}
\label{sec:method}
\subsection{Motion Map Representation}
\label{sub_sec:momap}
\begin{figure*}[t]
    \centering
    \includegraphics[width=1.0\linewidth]{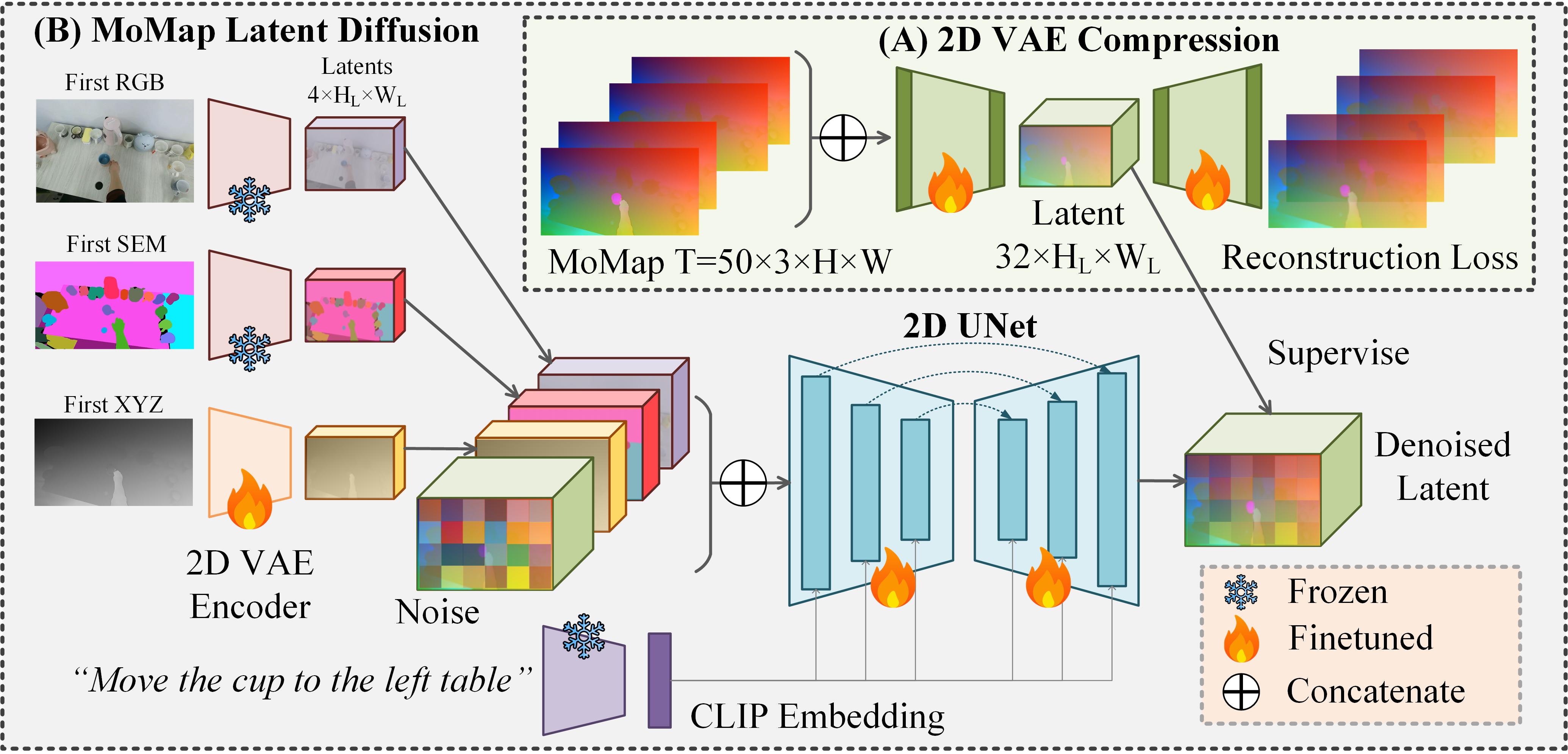}
    \caption{\small\textbf{Method Overview}: (A) A MoMap can be compressed to compact latent via initializing and finetuning a Stable-Diffusion VAE. (B) Given a starting frame and language condition, MoMap can be generated by finetuning the SD UNet.
    }
    \vspace{-2em}
    \label{fig:momap_main}
\end{figure*}
The first challenge is to represent real-world dynamic 3D scenes in a way that neural networks can easily output or predict. 
It is crucial to capture motion in 3D space rather than in 2D. 
Most prior methods model motion as 2D tracks---either for reconstruction tasks (e.g., BootsTAPIR) or for generating 2D tracks (e.g., TAPIR~\cite{doersch2023tapir}, ATM~\cite{wen2023any}, and Track2Act~\cite{bharadhwaj2024track2act}). 
However, 2D tracks require the model to handle complexities like occlusion.
Therefore, we aim to represent the dynamic scene directly as 3D trajectories.
A prior method, General-Flow~\cite{gflow}, attempts to model individual 3D trajectories as a set of polylines, using a PointNet-like structure to build a VAE that generates short-term 3D trajectories. 
However, this unstructured set representation cannot handle dense predictions for every pixel and has limited network capacity. 
Inspired by recent work repurposing image diffusion models for depth prediction (e.g., Marigold), we propose a novel scene representation, called Motion Maps, that is pixel-aligned and can be produced by a repurposed image diffusion model.
As shown in Fig.~\ref{fig:momap_repr}(A), a Motion Map (MoMap) is an image of 3D trajectories/curves, or intuitively, is a list of XYZ position maps aligned onto the single-view reference image. 
Given a reference time (e.g., $t = 1$), we store the 3D trajectory of each pixel observed at this reference time in the corresponding pixel location. 
All 3D locations are defined with respect to the fixed reference frame at this time, which separates the egocentric (camera) motion from scene object motion. 
Other ways to see it are that MoMap forms a ``3D trajectory image'' of all pixels anchored at the reference time; or that MoMap is an XYZ recoloring across time of the same reference XYZ point map.
The MoMap representation offers several advantages:
\begin{itemize}
    \item \textbf{Pixel-aligned}: MoMap shares the same underlying structure as a standard image, making it straightforward to generate using large 2D models and allowing us to leverage pretrained image priors.
    \item \textbf{Dense}: Unlike a set of sparse trajectories, MoMap is dense and stores information for every pixel at the reference time, thus capturing both background and foreground elements critical for understanding motion.
    \item \textbf{Smooth and Compressible}: Because MoMap stores 3D trajectories, which are often smooth and low-rank, it can be further compressed into compact latent maps.
    \item \textbf{Camera Motion Disentanglement}: By “freezing” the reference frame, MoMap isolates only the meaningful (foreground) motion, removing the camera’s egocentric motion.
\end{itemize}
Note that one MoMap only captures the content of pixels observed at the reference time, a more complete dynamic 3D scene can be represented by the union of multiple MoMaps anchored at different reference frames with camera poses. This paper only studies the generation of one MoMap and leaves the complete multi-MoMap joint generation as future work (Sec.~\ref{sec:conclusion}).

\subsection{MoMap Database from Real Videos}
A non-trivial question arises: where do we learn meaningful motion priors? Many previous works rely on synthetic data (e.g., Kubric and  PointOdyssey) to train 2D/3D tracking models that reconstruct trajectories from real observations, or they use manually/semi-automatically labeled small datasets (e.g., General-Flow) to generate short 3D trajectories. However, synthetic data often contains random, semantically/functionally meaningless motion that is only suitable for low-level vision tasks such as tracking or flow, while manual or semi-automatic labeling does not scale well to large video datasets. Instead, we leverage recent advances in 4D reconstruction (MoSca~\cite{lei2024mosca}) to reconstruct many real-world 4D scenes, and then convert them into our MoMap representation.

Specifically, as shown in Fig.~\ref{fig:momap_repr}(B), the data preparation stage processes all frames of a raw 2D real video with several steps:
\begin{enumerate}
    \item We first apply an off-the-shelf video depth model, DepthCrafter, to infer the depth of each frame. 
    \item We then query a 3D tracker (SpaTracker) densely on every pixel of a target reference time frame (randomly sampled for each video) to build the corresponding MoMap.
    \item For videos captured by a moving camera, we run a tracklet-based bundle adjustment to solve for camera poses.
    \item Because no current 3D tracker can reliably predict 3D locations under occlusions, we must run an optimization procedure to address occluded intervals in each 3D trajectory. We adapt the geometric optimization stage from MoSca~\cite{lei2024mosca} to optimize the dense foreground 3D trajectories.
    \item Finally, we fill in the MoMaps at the reference time using these optimized dense 3D trajectories.
    \item In addition to this geometric reconstruction process, we also apply a VOS model (DEVA~\cite{cheng2023tracking}) for dense instance-level video object segmentation, allowing us to incorporate semantic information into the later 3D motion generation.
\end{enumerate}
We applied this data processing pipeline to \textbf{3K HOI4D~\cite{liu2022hoi4d}} videos (human-object, egocentric camera) and \textbf{60K BRIDGE~\cite{walke2023bridgedata}} videos (robotics, static camera) as shown in Fig.~\ref{fig:momap_repr}(B).

\subsection{MoMap Generation}
\subsubsection{Task Definition}
\label{sec:momap_gen_task}
Given a large database of MoMaps, we aim to learn a prior for 3D motions. As shown in in Fig.~\ref{fig:teaser}(A), the task is as follows: \emph{given a single RGB image (the first frame) and optionally a text prompt, the model should generate the 3D dense motion of every pixel in that frame, from the input time step into the future time steps.}

We also aim to leverage semantic information from instance segmentation, since points belonging to the same instance typically share a similar motion pattern. Additionally, we assume access to a reliable monocular depth model so that the system does not need to learn 2D-to-3D lifting from scratch. Consequently, the first-frame RGB image can be augmented with an instance segmentation map and an XYZ map derived from the monocular depth model.

\subsubsection{MoMap Compression}\label{sec:momap_gen_compress}

Thanks to the pixel-aligned property of the MoMap representation, our 
approach is to exploit existing 2D image diffusion models (e.g., StableDiffusion) to generate MoMaps. However, these image models generally operate on inputs and outputs of shape $H\times W \times 3$, while MoMaps have shape $H\times W \times T \times 3$, where $T$ is the number of frames. 
Notably, $T$ is typically large (e.g., $50$ or $60$) in our setting, as we want to generate semantically and functionally meaningful long-term 3D motions. The key question, therefore, is how to adapt these powerful 2D diffusion models to generate MoMaps.
We found that it is necessary to encode the $H\times W \times 3$ MoMaps into compact latent feature maps of shape $H_L\times W_L \times C_L$. The underlying insight is that real-world motion is often smooth and low-rank, making it compressible in both space and time. Concretely, we set $H_L=H/8$ and $W_L=W/8$ for the spatial dimensions, and compress the temporal dimension $T=50$ (times the 3 spatial channels) into $C_L=32$ channels.
As illustrated in Fig.~\ref{fig:momap_main}, we increase the input, output, and latent dimensions of the 2D VAE. By carefully initializing the network at the input, intermediate, and output layers, we can effectively transfer weights from a pre-trained VAE, and then finetune with MoMap data. This design allows the architecture to function similarly to the standard 3-channel RGB (as averaging across time of XYZs), leveraging the pre-trained knowledge acquired from large-scale image datasets.

\subsubsection{MoMap Diffusion}\label{sec:momap_gen_unet}
We exploit the pixel-aligned nature of MoMap by repurposing a large pre-trained image generation model (e.g., Stable Diffusion) to generate these ``trajectory images''. 
Once the MoMap compression networks are trained, we can convert the ground-truth MoMap into a compact latent representation and use the U-Net to perform generation.
Concretely, as shown in Fig.~\ref{fig:momap_main}(B), we modify the input and output layers of a pre-trained U-Net from Stable Diffusion while initializing the remaining layers with weights from SD. The input conditions—the first-frame RGB image, the instance segmentation image (encoded as random RGB colors per patch), and the first-frame XYZ map—are all encoded by 2D VAE encoders into low-resolution latents, which are concatenated with noise and fed into the U-Net. Language prompts are incorporated in exactly the same manner as in standard Stable Diffusion.
Note that such a lightweight (equivalent to one single image generation) diffusion process can efficiently generate long ($T=50$) frames in a single sampling process, exploiting the pixel aligned nature of the MoMap representation.

\subsection{Application}
\begin{figure}[t]
    \centering
    \includegraphics[width=1.0\columnwidth]{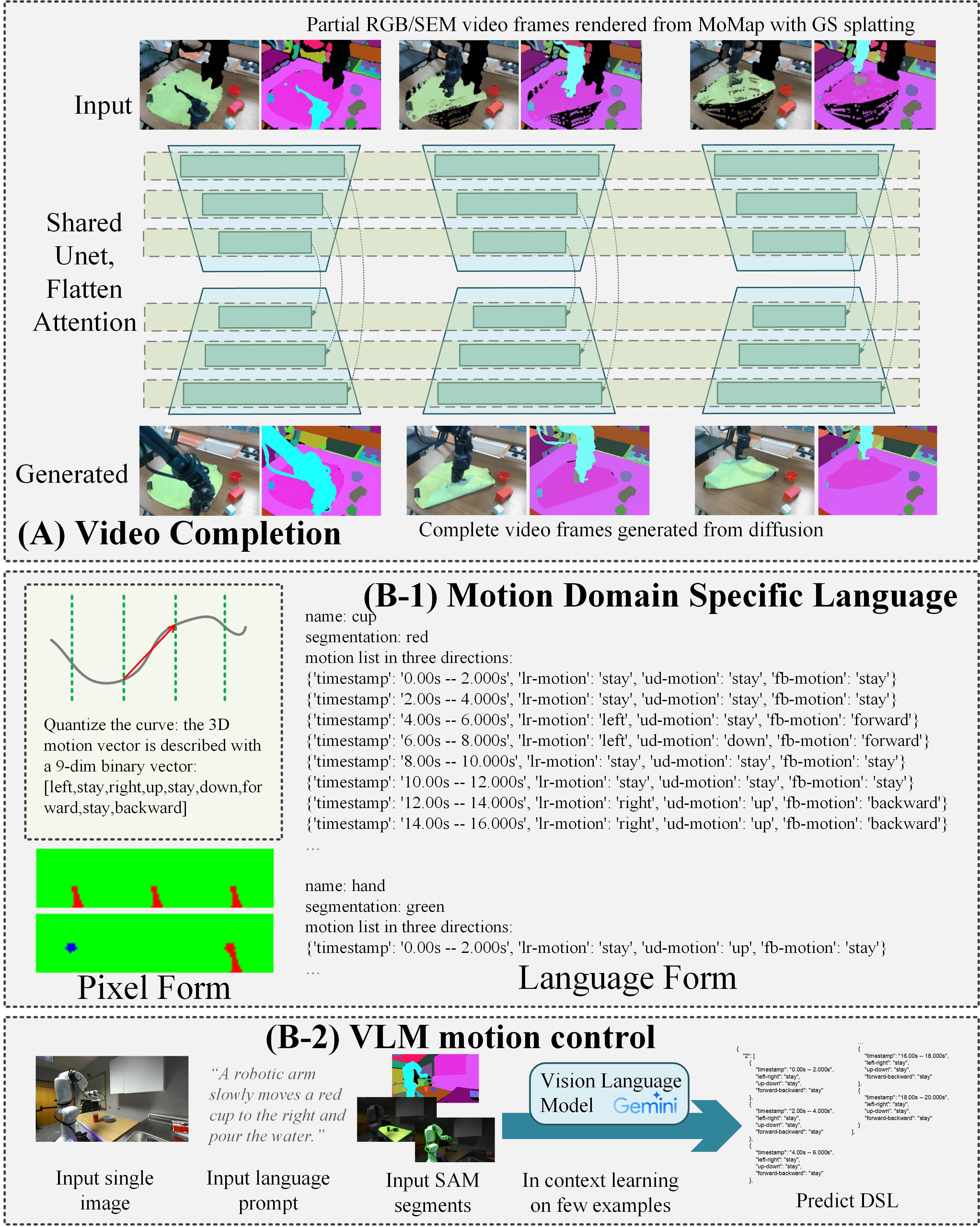}
    \caption{\small\textbf{Application}: 
    (A) 2D video generation via render MoMap and then complete; (B-1) motion DSL representation. (B-2) Infer motion DSL with VLM for finer generation control.
    }
    \vspace{-2em}
    \label{fig:momap_app}
\end{figure}

\subsubsection{Video Synthesis}\label{sec:method_app_video}
MoMap captures the future 3D motion of every pixel, which is inherently related to predicting future 2D video frames (commonly approached as a video generation problem). We further explore how to transform the 3D MoMap into a complete 2D video, as illustrated in Fig.~\ref{fig:momap_app}(A).
Given a generated MoMap and target camera intrinsics and extrinsics,
we render the MoMap using 3D Gaussian Splatting to obtain a partial 2D RGB and semantic video (top row). This rendering is partial because a single MoMap contains only the pixels visible in the first frame, resulting in holes (shown in black). We then finetune another Stable Diffusion single-image model to complete these partial videos using a simple cross-view flattening attention module (similar to CAT3D), ultimately reconstructing the full video (bottom row).

At a high level, this approach explicitly embeds a dynamic 3D inductive bias into video generation. Unlike most video models, which focus on memorizing pixel-level color changes and tend to be computationally heavy, we directly learn how objects move in 3D and form the images using a dedicated renderer (Gaussian Splatting). A separate diffusion model subsequently corrects artifacts and fills in missing regions. As a result, the video generation is handled by two relatively lightweight image models.

\subsubsection{VLM Control}\label{sec:method_app_mospeak}
Real-world motion often spans long time periods, making effective conditioning challenging. Relying solely on global text conditions (as in Stable Diffusion) may be insufficient and would require highly accurate text annotations in the dataset (Fig.~\ref{fig:exp_app}-B). Consequently, there is a need for alternative or augmented conditioning strategies to achieve robust and contextually aligned control over motion generation.
We propose leveraging Vision-Language Models (VLMs), which possess strong semantic understanding and commonsense reasoning. To connect high-level VLM outputs with low-level MoMap diffusion, we introduce a domain-specific language (DSL), illustrated in Fig.~\ref{fig:momap_app}-(B-1). The centroid of each semantic patch is quantized into nine directional flags (e.g., ``left'', ``right'', ``stay'') and converted into a structured language format (e.g., JSON) for each patch. Alternatively, these patch-level directional flags can be grounded at the pixel level, forming 2D conditions. Thus, this DSL seamlessly combines both a structured language format and pixel-grounded information.

Rather than inserting text prompts directly into the MoMap diffusion network, we remove the text condition and instead rely on pixel-based DSL (2D pixel conditions) to guide generation. 
During inference, we let a VLM (e.g., Gemini) generate the DSL from the input’s first frame as well as the global text prompt via in-context learning, using a few ground-truth examples from the training videos (Fig.~\ref{fig:momap_app}-B-2). In this setup, VLM's strong reasoning and language capabilities allow it to interpret flexible prompts—even those substantially different from the training dataset—and determine what moves and how. The DSL it produces is then converted into pixel-form conditions for the diffusion model’s U-Net (trained without global text conditioning) as an input condition, enabling fine-grained, semantically informed control over 3D motion generation.

\begin{figure*}[h]
    \centering
    \includegraphics[width=1.0\linewidth]{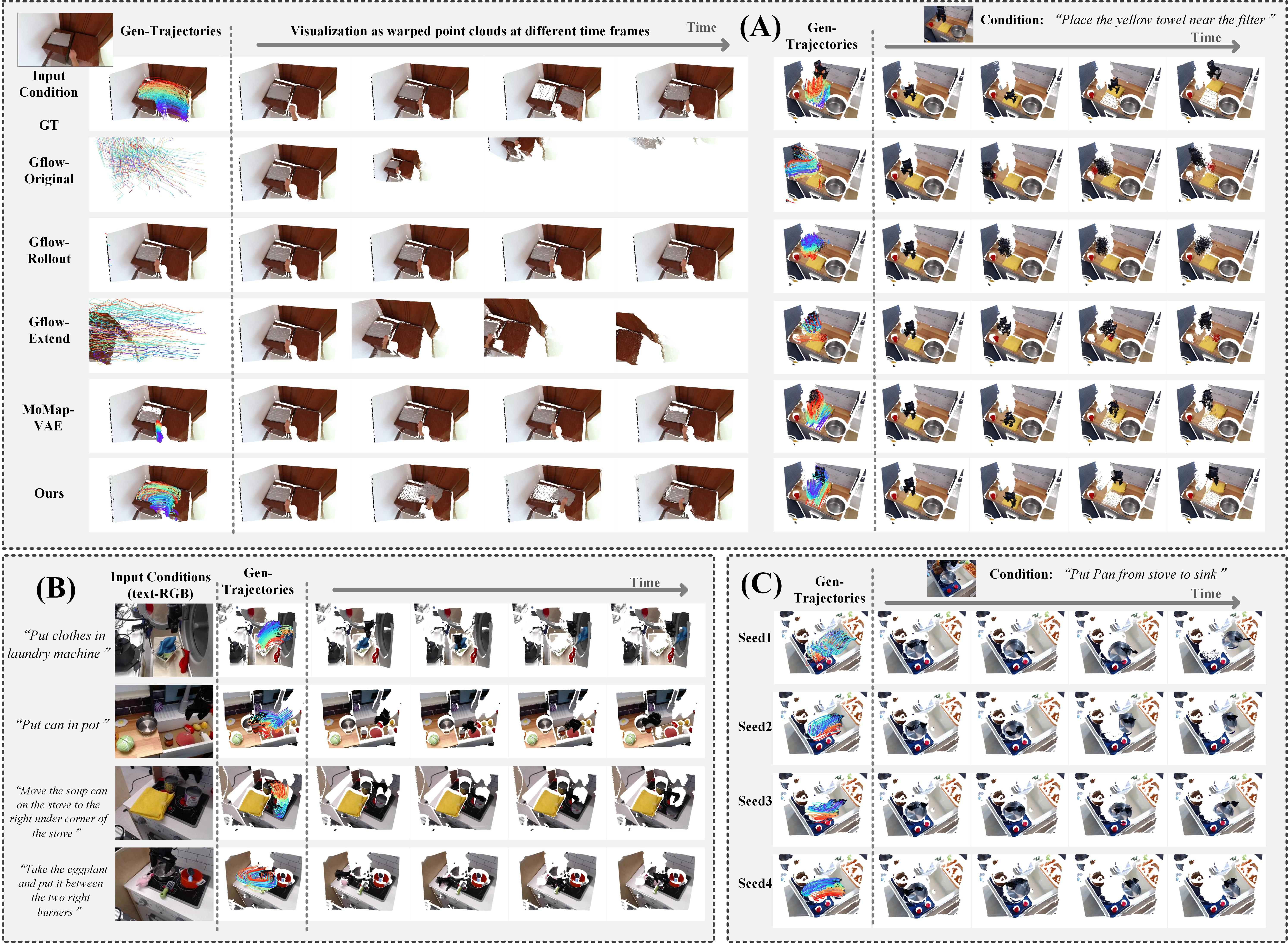}
    \caption{\small\textbf{Qualitative Results}: (A) comparison with baselines. (B) More generation results. (C) Diverse generations from the same condition with different random seeds.
    }
    \vspace{-2em}
    \label{fig:exp_comp}
\end{figure*}
\vspace{-2em}
\section{Experiment}
\label{sec:exp}

\subsection{Baselines Comparison}
\begin{figure}[t]
    \centering
    \includegraphics[width=1.0\columnwidth]{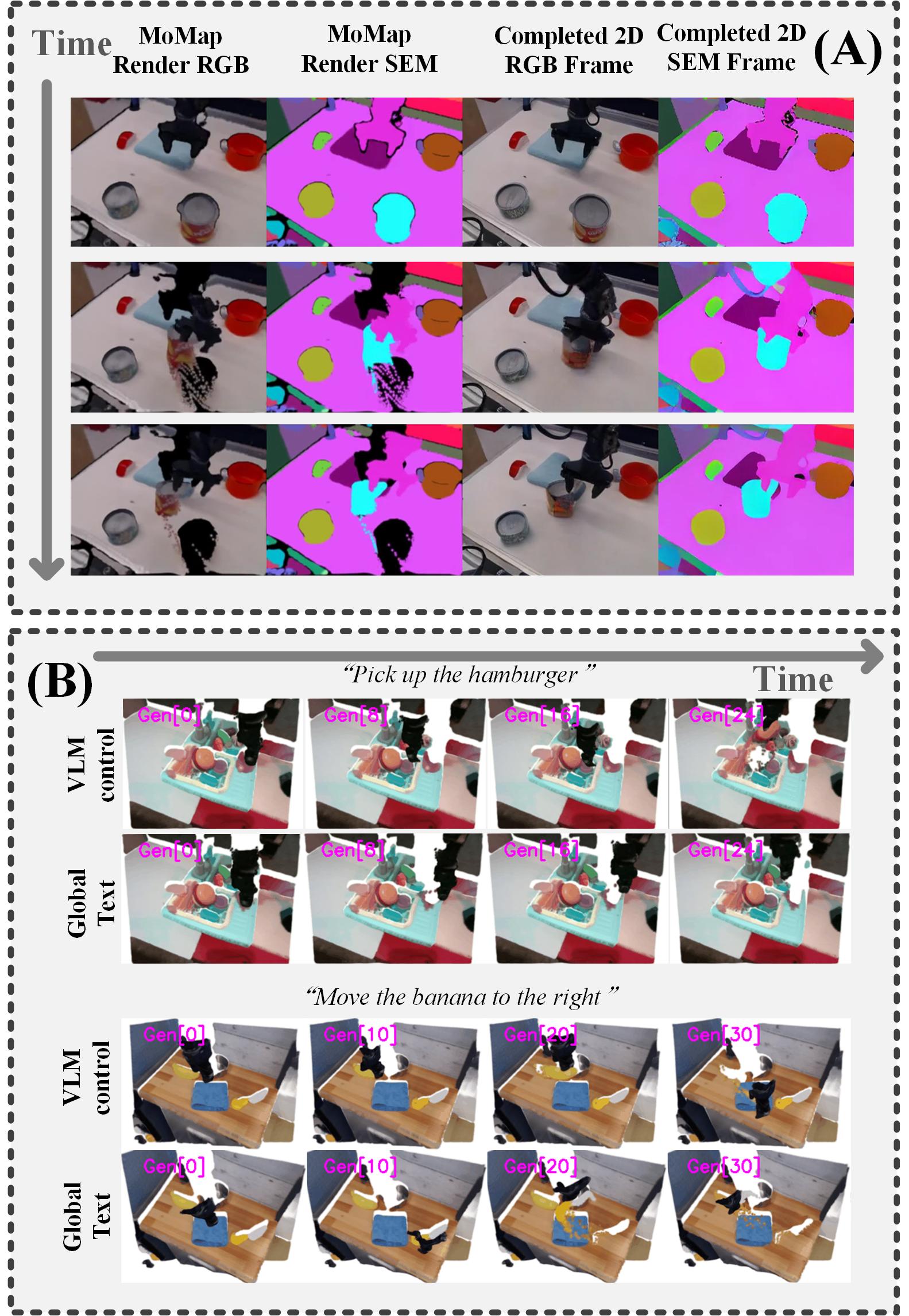}
    \caption{\small\textbf{Application Results}: (A) 2D video generation results. (B) VLM finer control can generate reasonable results even when global text condition fails.
    }
    \vspace{-2em}
    \label{fig:exp_app}
\end{figure}

We compare our 3D motion generation approach with several baselines that also generate 3D trajectories. The closest related work is GeneralFlow~\cite{gflow}, which produces short 3D trajectories (typically $T=4$ frames) from an input-colored point cloud. We train three variants of the G-Flow-large model on our BRIDGE robot dataset:
\begin{itemize}
    \item \textbf{GFlow-Original}: Uses the same configuration as the original GeneralFlow~\cite{gflow} but extends the trajectory prediction length from 4 to 50.
    \item \textbf{GFlow-Rollout}: Keeps the original configuration without changing the prediction length. To achieve T=50, it performs iterative ``roll-outs'' by taking the last generated frame as the new first-frame input for each subsequent generation round.
    \item \textbf{GFlow-Extend}: An upgraded version of GeneralFlow that increases the querying chunk size from 128 to 20,000 and the scene point cloud samples from 4,096 to 10,000. The network also takes semantic colors as additional point features. Furthermore, the trajectory VAE latent dimension is increased from 16 to 32, and more layers are added to boost overall capacity.
\end{itemize}
We also built another baseline, \textbf{MoMap-VAE}, which uses the same overall architecture but omits the diffusion U-Net in the middle. Instead, it is trained as a conditional VAE that directly generates the MoMap.

Generating a long motion (T=50) from a single starting frame is highly ill-posed, so the most direct and arguably maybe the most accurate way to assess performance is via side-by-side qualitative comparisons of the generated 3D motions as in Fig.~\ref{fig:exp_comp}.
From the comparisons, we observe:
\textbf{GFlow-Original}: Tends to produce diverging trajectories in long-term predictions. Objects often lose their original rigid shape at later time steps.
\textbf{GFlow-Rollout}: Performs poorly in this setting because it can only generate very short time spans (T=4) and must roll out repeatedly to reach T=50. This causes even more severe divergence over time.
\textbf{GFlow-Extend}: Benefiting from a larger query chunk size and more network capacity, it generates smoother and more consistent motion. However, due to limited semantic and contextual understanding of Point-NeXT, it misidentifies which objects should move. For instance, in the second-to-last row, it moves two blocks when only one should be moving.
\textbf{MoMap-VAE} vs. \textbf{Ours} (MoMap Diffusion): While MoMap-VAE can produce reasonable long-term motions, our diffusion-based approach generates more natural-looking hand and object motions, reflecting a richer learned prior over realistic 3D trajectories.

\subsection{Metrics and Quantitative Comparison}

Evaluating the generated results quantitatively is challenging because the problem is highly ill-posed and many plausible generations can be consistent with the provided initial conditions. For each starting frame, there is only one ground-truth observed trajectory, but many plausible solutions exist. We follow the previous work (GeneralFlow~\cite{gflow}) which generates N = 10 samples per input, but computing the metric by comparing the \textbf{closest} of the generated samples with the ground truth. 
Beside detailed reconstruction erros, we introduce several additional coarse metrics that capture various aspects of the motion:

\begin{enumerate}
    \item Foreground Mask IoU (\textbf{fg\_mask\_iou $\uparrow$}): Measures whether the generated motion correctly identifies the moving regions. A ``moving mas'' is constructed by identifying pixels with significant 3D trajectory displacement. The Intersection over Union (IoU) between the ground truth and generated moving masks is then computed.
    \item Foreground Absolute Trajectory Error under Dynamic Time Warping (\textbf{ate\_dtw$\downarrow$}): Computes ATE (reported in GeneralFlow) after optimizing temporal alignment with Dynamic-Time-Warping (DTW), capturing scenarios where the timing of motions differs but the overall trajectory is correct.
    \item Foreground Curve Distance-Matrix Signature Error (\textbf{D\_sig$\downarrow$}): Creates a translation and rotation-invariant signature by computing a T x T distance matrix between positions at any two-time steps. The difference between ground truth and generated matrices measures SE(3) invariant curve similarity.
    \item Foreground Local Distance Difference \textbf{(local\_dist\_diff$\downarrow$}): Evaluates the preservation of local structure by computing distances between K-nearest neighbors in the ground truth and generated 3D trajectories. Measures whether the generated motion preserves local rigidity and avoids collapsing into noise.
    \item Foreground Patches Nearest Patch Accuracy (\textbf{cross\_patch\_nearest\_acc$\downarrow$}) tracks which patch is closest (centroid-wise) to each moving foreground patch at each time step. Accuracy measures whether the generated motion maintains the same nearest-patch relationships as the ground truth.
    \item Foreground dT Quantized Direction Accuracy (\textbf{quantize\_acc\_dT$\uparrow$}): Coarsely evaluates motion direction by quantizing a 3D trajectory over a delta time interval (dT) into 9 directions (left-stay-right, forward-stay-backward, up-stay-down). Measures the accuracy between ground truth and generated motion directions for each pixel in the foreground mask.
\end{enumerate}
These metrics capture a variety of aspects, ranging from fine-grained reconstruction errors to coarse object-level motion directions and proximities, enabling a more comprehensive evaluation of 3D trajectories. The metrics of Ours and the baselines on a small testset are reported in Tab.~\ref{tab:exp_comp}.
\begin{table*}[t]
\centering
\scalebox{0.7}{
\begin{tabular}{@{}|c|ccccc|ccccc|@{}}
\toprule
Dataset                       & \multicolumn{5}{c|}{{\color[HTML]{181A1B} BRIDGE}}                                                & \multicolumn{5}{c|}{{\color[HTML]{181A1B} HOI4D}}                                                 \\ \midrule
Metric\textbackslash{}Method  & \textbf{GFlow-Ori.} & \textbf{GFlow-Roll.} & \textbf{GFlow-Ext.} & \textbf{MoMap VAE} & \textbf{Ours}   & \textbf{GFlow-Ori.} & \textbf{GFlow-Roll.} & \textbf{GFlow-Ext.} & \textbf{MoMap VAE} & \textbf{Ours}   \\ \midrule
\textbf{fg\_mask\_iou↑}       & 0.8044             & 0.5659            & 0.7597            & \textbf{0.8271}    & {\ul 0.8128}    & 0.0667             & 0.0000            & 0.0667            & {\ul 0.3400}       & \textbf{0.4492} \\
\textbf{ate\_dtw↓}            & {\ul 0.0747}       & 0.1164            & 0.0812            & 0.0751             & \textbf{0.0689} & 1.6119             & {\ul 0.1241}      & 0.7281            & 0.1265             & \textbf{0.1112} \\
\textbf{D\_sig↓}              & {\ul 0.0536}       & 0.0873            & 0.0591            & 0.0538             & \textbf{0.0463} & 1.3888             & 0.1193            & 0.2922            & {\ul 0.1030}       & \textbf{0.0886} \\
\textbf{local\_dist\_diff↓}   & 0.0124             & 0.0154            & 0.0073            & {\ul 0.0073}       & \textbf{0.0058} & 0.0154             & \textbf{0.0086}   & {\ul 0.0092}      & 0.0094             & {\ul 0.0092}    \\
\textbf{patch\_nearest\_acc↑} & 0.8387             & 0.7649            & 0.8244            & {\ul 0.8398}       & \textbf{0.8691} & 0.6926             & {\ul 0.7631}      & 0.7554            & 0.7495             & \textbf{0.7923} \\
\textbf{quantize\_acc\_1↑}    & 0.6758             & 0.6271            & 0.6623            & {\ul 0.6856}       & \textbf{0.7212} & 0.2116             & 0.6014            & 0.3279            & {\ul 0.6015}       & \textbf{0.6157} \\
\textbf{quantize\_acc\_4↑}    & 0.5890             & 0.4702            & 0.5568            & {\ul 0.6148}       & \textbf{0.6554} & 0.3653             & 0.2710            & {\ul 0.3656}      & 0.3000             & \textbf{0.4278} \\
\textbf{quantize\_acc\_16↑}   & {\ul 0.7294}       & 0.4503            & 0.7049            & 0.7293             & \textbf{0.7752} & {\ul 0.4319}       & 0.1347            & 0.4068            & 0.3980             & \textbf{0.4610} \\ \bottomrule
\end{tabular}
}
\caption{Quantitative comparison on BRIDGE and HOI4D testset.}
\vspace{-1em}
\label{tab:exp_comp}
\end{table*}

We observe that our method outperforms the baselines in most metrics, especially in quantized directional accuracy and the D\_sig\_fg$\downarrow$ metric. 
From both qualitative and quantitative findings, we note that our generated trajectories exhibit stronger semantic coherence and greater internal consistency, particularly in capturing aggregate motion. In contrast, GFlow appears to have lower noise in individual point trajectories (reflected by higher reconstruction metrics) but exhibits inconsistent noise across adjacent points belonging to the same object, leading to visually detectable ``object breakdowns.''

\subsection{Application}
\paragraph{Video synthesis.} We present our video synthesis results in Fig.\ref{fig:exp_app}-A, following the approach in Sec.\ref{sec:method_app_video}, showing that our lightweight model achieves decent performance.
\vspace{-1em}
\paragraph{VLM control.} In Fig.\ref{fig:exp_app}-B, we showcase VLM-controlled generation, described in Sec.\ref{sec:method_app_mospeak}. Notably, the VLM-controlled approach produces more accurate motion than relying solely on a global text condition, which can fail in certain scenarios.

\subsection{Ablation Study}
We validate the design of our MoMap diffusion by ablating various components in its architecture: (1) remove the input XYZ condition; (2) remove the input semantic image condition; (3) remove the VAE compression of the MoMaps and let the UNet work on the original MoMap space instead of the latent space; (4) train the diffusion from scratch without initialization from stable-diffusion pre-trained weights; and (5) train both the MoMap VAE and UNet from scratch. Tab.~\ref{tab:exp_abl} presents the results of these ablation experiments, illustrating that our full model achieves the best performance.

\begin{table}[t]
\scalebox{0.6}{
\begin{tabular}{@{}ccccccc@{}}
\toprule
Metric\textbackslash{}Ablation & Full            & no xyz & no sem          & no vae & no unet init    & no unet\&vae init \\ \midrule
\textbf{fg\_mask\_iou↑}        & \textbf{0.8128} & 0.7962 & 0.8079          & 0.1305 & 0.7974          & 0.7450            \\
\textbf{ate\_dtw↓}             & \textbf{0.0689} & 0.0738 & \textbf{0.0689} & 0.1759 & 0.0792          & 0.0992            \\
\textbf{D\_sig↓}               & \textbf{0.0463} & 0.0484 & 0.0468          & 0.1082 & 0.0520          & 0.0579            \\
\textbf{local\_dist\_diff↓}    & 0.0058          & 0.0060 & 0.0056          & 0.1348 & \textbf{0.0055} & 0.0094            \\
\textbf{patch\_nearest\_acc↑}  & \textbf{0.8691} & 0.8605 & 0.8659          & 0.7533 & 0.8510          & 0.8306            \\
\textbf{quantize\_acc\_1↑}     & \textbf{0.7212} & 0.7142 & 0.7187          & 0.1940 & 0.7085          & 0.6566            \\
\textbf{quantize\_acc\_4↑}     & \textbf{0.6554} & 0.6415 & 0.6522          & 0.3977 & 0.6260          & 0.5925            \\
\textbf{quantize\_acc\_16↑}    & \textbf{0.7752} & 0.7589 & 0.7719          & 0.5311 & 0.7434          & 0.7267            \\ \bottomrule
\end{tabular}
}
\caption{Ablation comparison on BRIDGE dataset.}
\label{tab:exp_abl}
\end{table}

\section{Limitations and Conclusion} \label{sec:conclusion}

In this paper, we explored a new problem: learning to generate future 3D motion for an entire scene using large-scale real-world videos. We introduced an image-like \emph{MoMap} representation, which enables the repurposing of established 2D image diffusion models for 3D motion generation. By assembling a large database of MoMaps from real videos, we demonstrated the feasibility of synthesizing semantically and functionally meaningful 3D scene motion, also highlighting its potential impact on  2D video synthesis.

Despite these promising results, our work represents an initial step in the realm of dense 3D motion generation at scale, and several challenges and directions remain:

\begin{enumerate} 
\item \textbf{Multiview MoMap joint generation.}
This paper primarily focuses on generating a single MoMap anchored at the first frame, which only partially captures the scene. A key future direction lies in jointly and consistently generating multiple MoMaps anchored at different frames or viewpoints to aggregate a more complete scene.

\item \textbf{Enhanced motion control and VLMs.}  
Currently, the language prompts are integrated similarly to Stable Diffusion, providing limited motion control. Achieving more fine-grained and semantically meaningful motion control using advanced vision-language models is a promising area of future research.

\item \textbf{Scaling to larger and more diverse data.}  
Our current datasets, though sizable, are domain-specific (e.g., hand-object or robotics). Extending these techniques to large-scale, general-purpose videos presents an important avenue for broader impact.
\end{enumerate}

\newpage
{
    \small
    \bibliographystyle{ieeenat_fullname}
    \bibliography{main}
}

\end{document}